# The Practimum-Optimum™ Algorithm for Manufacturing Scheduling: A Paradigm Shift Leading to Breakthroughs in Scale and Performance

**Moshe BenBassat, Chairman of Plataine Inc. (moshe.benbassat@plataine.com)**
If you wish to read more about my AI work and earlier publications, please visit www.moshebenbassat.com, contact me at: moshe.benbassat@plataine.com

## 1. INTRODUCTION

Generating an optimal production schedule is essential for manufacturing operations, having a significant impact on the efficiency and productivity of the whole operation and, hence, on top and bottom line, as well as on-time delivery dates. Yet, building a high-quality schedule for large complex operations has been known to be one of the most challenging business tasks. Additionally, such large-scale automation enables the automatic generation of optimal schedules for a long time horizon, e.g., months in advance. This opens the door to new AI applications for promising delivery dates based on a more accurate assessment of future resource availability, and for tactical capacity planning.

## 2. Optimizing Manufacturing Schedule: The Problem

At the end of a regular weekly or monthly cycle, the scheduler in a manufacturing company faces a Demand Set (Fig. 2 below) of customer work orders (WO) that need to be scheduled for production over the coming time horizon (next week, next month, or longer). For simplicity, we assume that each work order specifies a single product, the required quantity, and the expected delivery date. Typically, the production process of each product consists of multiple tasks (steps) that need to be executed in a particular order, with each task requiring specific machine(s), tools, raw materials, and people. There may also be dependencies between work orders.

The scheduler's job is to build a schedule at the task level that optimizes an objective function Z representing a mix of schedule quality goals, such as due-date compliance, makespan (time of finishing the last task), total throughput, raw material consumption, machine utilization, and others. These are sometimes assigned weights that enable consolidating them into a weighted Z score for the overall schedule quality. For example, in the case illustrated in Figure 1 below, Due Date Compliance was set as the top-ranked goal relative to Autoclave Utilization and Other Stations Utilization

The schedule is graphically represented by the well-known Gantt board (see Figure 1), where each tile on the scheduling board represents an **assignment** that specifies the production step for one of the products, its start and end times, and all the resources it requires: material, machines, tools, and people. All entities are associated and connected with **operational requirements and constraints** rules that establish the criteria for valid assignments on the Gantt board. Let's elaborate on this important point.

**Figure 1- A Gantt Board**

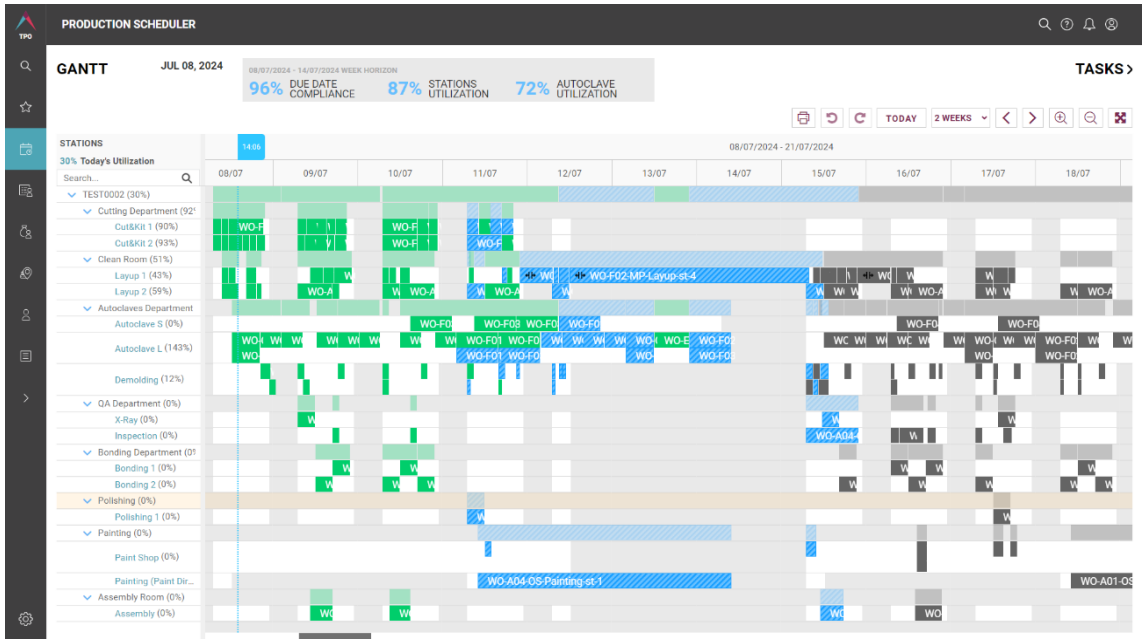

## 3. Requirements and Constraints are the Rules of the Game

Some of the **requirements and constraints** are obvious 'technical' rules, e.g., you cannot cut a 60" wide roll on a 40" wide machine.

Others relate to standard working hours (shifts) and availability of each resource with a rule that states that an assignment's time must be within the working hours of all its resources. Some machines work night shifts, while other resources are only available during one or two shifts.

Another example: clearly, assignments must comply with the sequence in which tasks are to be performed. Yet, for some tasks, it is allowed that Task B may start after Task A reaches a given percentage of its completion time.

Some machines, such as autoclaves, can simultaneously process multiple parts. Such task grouping must be consistent with batching rules (which parts can go together for a joint session).

Another example is a spreading & cutting station where multiple people can jointly work. In such cases, there will be a rule of "not less than" and "not more than," and what can be assumed as for the productivity of a given number of people (e.g., is it twice as fast for two people, or just 1.5x faster)

Large-scale complex operations may involve tens of production machines, multiple tool types, a variety of components and raw materials, and, of course, many people, with each operator having their own skills that qualify them as a candidate to operate certain machines. This entails a long list of requirements and constraints rules that need to be checked before an assignment can be declared a VALID candidate at a certain point and incorporated in the search process on the way to building an optimal schedule.

Requirements and constraints are the **Rules of the Game**. If a scheduling software cannot play by these rules, in other words, if it could generate invalid schedules, then there is no point in evaluating the quality of its schedules. Violating the rules of the game is a showstopper. Hence a prerequisite from any commercial scheduling software for a specific industry vertical is the ability to express all validity rules with their true practical meaning and incorporate them in the algorithm for placing tiles on the Gantt chart schedule. It is true that **some** rules may be flexible, but most are **MUST** rules. Each industry vertical (e.g., aerospace, automotive, semiconductors) and each specific factory has its own unique products and its own production processes, machines, tools, components, and raw materials. Accordingly, it has its own constraints and requirements that establish the 'rules of the game' criteria for valid assignments on the schedule. Getting to the point where software plays by the rules of the game requires deep domain expertise in the target industry vertical.

## 4. Current Practice Algorithms

Many algorithms have been proposed to automatically generate production schedules for complex manufacturing operations. Some examples are:

- Simulated annealing, e.g. [1]
- Tabu search, e.g. [2]
- Genetic algorithms, e.g. [6]
- Grasshopper algorithms, e.g. [5].

As well there are heuristic [4] search algorithms that are based on principles used by human expert schedulers, such as:

- Backward scheduling
- Forward scheduling
- Bottleneck-oriented scheduling.

We do not include here algorithms based on 'exact' mathematical models, such as Integer Programming because they don't handle well the requirements and constraints of complex operations that define valid schedules and thus are not applicable, to begin with. This was first covered in our 1990 article: "The **W-6** Approach to Multi-Dimensional Scheduling: Where AI and Operations Research meet."

Most algorithms in current use are based on a strategy that, by analogy, roughly follows the hill-climbing process. From an initial start point of an empty Gantt board, such an algorithm uses systematic step-by-step logic for selecting and placing the next tile on the Gantt board. At each step, the algorithm assesses the Gantt board's quality status and looks for steps that will take it toward a higher-quality schedule.

When it reaches a point where adding or modifying tiles does not improve the schedule quality (i.e., a local peak), the algorithm proceeds to look for options to move away from the local peak position. These include making local micro changes to existing tiles on the board, or even removing some tiles from the board, in an attempt to find a new starting point from where it can restart the process of climbing up in the direction of peaks that are higher than those previously attained. When it reaches a schedule judged to be the best (highest peak), or it runs out of time, it stops and presents to the user the best schedule it ended with.

Each algorithm in current practice has its own 'algorithmic logic' for (a) making the next small steps up and (b) how to recover from a 'local peak' schedule. It iteratively repeats its own specific logic with the objective of reaching the top peak with a minimal number of iterations. The process is based on small

(micro) moves on the path towards the best schedule. It is worth noting that the majority of contemporary algorithms spend most of the computing time exploring only a small part of the gigantic universe of all possible valid schedules.

In the end, they arrive at a single schedule presented as the best or 'optimal.' As for scale, when it comes to scheduling complex manufacturing operations, commercially available software products max out at 500-1,000 tasks in fully automatic mode. Another weakness of some of the algorithms is that the tiles' arrangement on the Gantt of the best schedule that is presented to the user is not 'intuitive' to human logic. This introduces difficulties in making changes when disruptions take place during schedule execution.

## 5. The New Practimum-Optimum™ Algorithm: An Overview

The limitations of the algorithms in the current generation of manufacturing scheduling software were the main drivers for the development of the new P-O algorithm described in this article with the following unique differentiators:

1. P-O routinely scales up to automatically schedule tens of thousands of jobs for complex manufacturing environments. See below in Section 9 specific real-life cases.
2. It explores broader areas of the universe of valid Gantt boards and high-quality schedules.
3. Its process includes a self-generating dataset of schedules, which it uses for its RL (Reinforced) machine learning.
4. At the end of each run, it generates multiple high-quality schedules with a variety of 'flavors' in terms of the mix of the quality metrics and the arrangements of the tiles on the Gantt board. (In contrast to all other algorithms that generate a single schedule at the end of each run). These are stored and later used in deciding/composing the best schedule presented to the user and/or making adjustments to it in response to the user's comments on the best schedule. In fact, quite often, schedules with different tile arrangements on the Gantt board have very close or the same scores of quality metrics.
5. P-O generates 'human-oriented' schedules, in other words, schedules that make business sense to a person.

The following sections provide more details on the P-O algorithm.

## 6. Factory's Digital Twins

The software infrastructure of the P-O algorithm is based on a complete digital representation of the manufacturing environment. This includes digital twins for all real physical objects, e.g., products, machines, tools, raw materials, people, and abstract objects such as Work Orders, Working Shifts, Assignments, and Gantt Charts. With these as the basis, all requirements and constraints are expressed in software structures to capture all the delicate real-life details that may be required for decision-making.

Plataine's Scheduler includes a configuration module that generates all the elements required for a full problem statement of a manufacturing scheduling problem in a given organization, e.g., products, machines, raw material, …, requirements and constraints,…etc. Most of these elements are fairly permanent. All that is left for a specific schedule-building session is to specify the Demand Set and any necessary revisions, such as, for example, any modifications in resource availability (machine M1 is on preventive maintenance this Thursday).

The Plataine Production Scheduler is one product in a suite of enterprise AI applications for decision optimization in manufacturing operations. This enterprise-level AI suite operates as an AI layer on top of existing ERP/MES data processing software. It includes several integrated AI-based decision optimization applications for the connected factory of the future. With its built-in APIs, the suite accesses any data that exists in ERP/MES systems and updates these systems with the data and decisions it generates. In the next sections, we assume that a factory site has already been configured with the digital twins for the P-O algorithm.

## 7. The Practimum-Optimum™ Algorithm: Main Components

The four main components of the P-O algorithm are:

### 7.1 Virtual Human Experts

Central to P-O is the introduction of a unique concept, the **Virtual Human Expert** (VHE) agent. Essentially, a VHE is an algorithm that operates with some 'school of thought' and heuristics that human experts use to build "optimal" schedules. These methods are described in textbooks, industry-related articles, and training classes for schedulers. Such methods/heuristics include, for example, forward scheduling, backward scheduling, shifting bottleneck heuristics, and variants or combinations of them, including switching the focus and priorities between products, machines, raw materials, and time. In reality, there is no one dominant methodology that fits all types of manufacturing operations in all industry verticals. Even for a given factory, some variations in the Demand Set to be scheduled, or temporal variations in machines' capacities, may cause Method A to produce better schedules than Method B, while, for another variation, Method B will be better. That is why in P-O, we selected the VHEs to represent a wide spectrum of variants of good human methods and heuristics and accordingly implemented in software the scheduling logic of a variety of VHEs.

### 7.2 A Single Schedule Building Iteration of a VHE Algorithm

Each VHE starts with a first iteration of applying its own scheduling logic to build a full schedule for the given Demand Set. That is, in a stepwise manner, it creates tiles on the Gantt board as determined by its specific VHE logic (in the same way that a human expert scheduler would) until it exhausts all jobs in the Demand Set or reaches a stopping rule.

The end result of each iteration (a pass through the full set of jobs in the Demand Set) is a valid Gantt board schedule. It could be that some jobs in the Demand Set did not fit and were left out of the Gantt board, as could happen in any algorithm.

An iteration does not include micro moves to attempt to improve the schedule Gantt board at the end of a pass. This is done by a reinforced (RL) machine learning algorithm, as described below.

### 7.3 Schedule Metrics

Each VHE calculates a rich set of tens of metrics covering macro metrics and micro metrics for each schedule it generates:

- **Macro** metrics are calculated over the entire Gantt board. For example, total throughput, overall percentage of due date compliance, overall consumption of raw material, or overall machine utilization.

- **Micro** metrics deal with selected sets of one or more assignments of the schedule, or for specific resources across the entire schedule, such as utilization of a designated resource, or due date performance of specific high-priority jobs.

The macro and micro metrics become part of every stored schedule and play a key role in the machine-learning processes within a single run and for longer-term learning.

## 7.4   Reinforced Learning (RL) Algorithm.

The orientation of the following machine learning algorithm is **not** to identify **micro moves** that will improve tile arrangement around a local peak. Rather, taking as input a fully valid Gantt Board and its rich set of metrics, the RL's goal is to learn sources of strengths and weaknesses of the schedule state. Accordingly, derive 'reward and punishment' changes in the Demand Set that will modify the relative priorities for time and resource allocation that jobs received in the prior iteration that led to the current state of the schedule.

As an example, assume an iteration by VHE3 whose logic is driven by the heuristic: "Due date jobs go first." With this heuristic, Job208 was the 2nd tile to be placed while Job26 was the 123rd to start its first task out of eight, some of them quite long. As a result, Job26 was left out of the schedule. On the other hand, for Job208 all tasks are relatively short, and it finished way before its due date. So, we detected an inadequacy. Clearly, Job208 had more degrees of freedom for time and resources to select from, compared to Job26, and we "spoiled" it a little, resulting in it finishing way ahead of its due date. On the other hand, Job26 received less than it deserved ("deprived"). If in the next iteration of VHE3, we give Job208 a small "reward," causing it to start earlier, and give Job26 a small "punishment," causing it to start a little later but still finish on its due date, there is a good chance the above inadequacy will not appear. For simplicity, in this example, we only used two jobs and one type of inadequacy. In fact, the P-O RL algorithms analyze the entire schedule at the end of every given iteration.

The post-learning changes of the RL machine learning translate into parameters in the jobs of the Demand Set that impact 'select next' and 'place next' decisions in the next iteration executed by the VHE. These cause the core logic of the VHE algorithm to explore, in the subsequent iteration, substantially different parts of the scheduling universe and potentially find higher-quality schedules. In most cases, this leads to radically different tile arrangements on the Gantt board and higher-quality schedules as iterations progress. Using the hill climbing analogy, this may be viewed as a big backtracking jump, shifting from a given local peak to a faraway promising start point. This is a fundamental difference from most contemporary algorithms, which spend considerable time in micro local steps restricted to the neighborhoods of local peaks they visit.

## Expanding the Analogy: Add Paragliding to Foot Climbing

Let us extend the mountain climbing analogy by adding parachute gliding capability to the existing climbing methods. Looking around from a local peak (P1), and recognizing the limitations of your current schedule, you paraglide - rather than walk - to another starting point from which there is good reason to expect that it could take you to a higher peak (P2). Gliding takes just a few minutes compared to the long hours it would take to walk down from P1 and reach the next start point. This analogy is a good way to explain a fundamental advantage of the Practimum Optimum algorithm. The phrase above: 'good reasons to expect…' is achieved by the RL learning mechanism

The next section describes how these four main components of P-O are integrated to produce the best schedule(s) for a given Demand Set.

# 8. Flow of the Practimum-Optimum™ Algorithm for Building an Optimal Schedule

As discussed above, the P-O algorithm includes multiple independent digital virtual human expert agents (VHEs) representing a variety of 'schools of thought' and heuristics for building optimal schedules. To produce an optimal schedule for a given Demand Set, each VHE performs multiple passes as follows:

## Building Gantt Boards with Multiple Independent VHEs

Given a Demand Set as input (Fig 2), and an initial starting Gantt board (mostly empty with possibly some residual tiles from last week), each VHE runs an iteration aiming to build a full valid schedule. Each schedule that is output by the run is illustrated below with the designation **SVHE(k,j)** where k indicates the VHE identity, and j indicates the iteration.

**Figure 2- Part of a Demand Set**

| O Name | WO Quantity | End Product | WO Due Date | Priority |
|---|---|---|---|---|
| WO-1001 | 200 | KT-1-1001 | 10/09/2024 | High |
| WO-1002 | 160 | KT-1-1002 | 10/09/2024 | High |
| WO-1003 | 1 | KT-1-1003 | 11/09/2024 | High |
| WO-1004 | 5 | KT-1-1004 | 12/09/2024 | Regular |
| WO-1005 | 10 | KT-1-1005 | 27/09/2024 | Regular |
| WO-1006 | 15 | KT-1-1006 | 14/09/2024 | Low |
| WO-1007 | 35 | KT-1-1007 | 27/09/2024 | Regular |
| WO-1008 | 53 | KT-1-1008 | 27/09/2024 | Regular |

Iteration j=1 by VHE k

All jobs of the Demand Set are placed on the Gantt board (just like a human expert would do, except much faster) and produce a **valid** schedule—**SVHK(k,1).** Valid means it complies with all constraints and requirements. In addition to the digital Gantt chart, the VHE also calculates a rich set of macro and micro metrics.

The tiles arrangement of this schedule SVHK(k,1), as well as its metrics, **are then submitted to the Practimum-Optimum Reinforced Machine Learning (RL) algorithm**, which produces updates to parameters in the Demand Set for the next iteration of VHE(k).

Iteration j=2 by VHE k

With the results of the RL machine learning phase embedded in the Demand Set at the end of prior iteration, the next schedule-building iteration starts building a new Gantt board for the <u>same</u> set of jobs. It ends up with a new **valid schedule SVHE(k,2),** which, due to the learning, is expected but not guaranteed to have higher quality than SVHE(k,1). RL machine learning is now applied to **SVHE(k,2),** which produces updates to parameters in the Demand Set for the next iteration of VHE(k).

…

…

Iteration j= n

VHE k continues this iterative process until a stopping rule is satisfied for VHE(k) (or it reaches its preset run time), and its **schedule SVHE(k,n)** is recorded as such.

Building the Schedules Dataset (**SDS**)

All of the schedules generated during this process by <u>all</u> VHE(k) agents are stored (digital Gantt of tiles arrangement and metrics) in an SDS repository for the given Demand Set. (In Machine Learning terminology, the macro and micro metrics may be viewed as the **features** of a schedule).

Announcing the Best Schedule

After all VHE agents (as many as exist) independently complete the above process, P-O searches the SDS repository for the best schedule in terms of the pre-set objective function and presents it to the user as the initial best schedule (**IBS**) that optimizes the pre-set macro goals while satisfying all requirements and constraints. In addition to the score values of each macro goal, the IBS also includes a list of insights that call the user's attention to important points about the IBS schedule (**Fig 3**).

**Figure 3- List of Insights is Part of Any Schedule**

| Task Name | Work Order Name | Insight Message |
|---|---|---|
| WO-1004 - Layup | WO-1004 | Task not scheduled due to shortage in layup stations. |
| WO-1018- Layup | WO-1018 | Task not scheduled due to lack shortage in layup stations. |
| WO-1017- Cutting | WO-1017 | Work order duration beyond the time horizon |
| WO-1019- Layup | WO-1019 | Task not scheduled due to shortage in "moldX" tools. |

As the user reviews the IBS and wishes to make further local changes, all they have to do is state WHAT they wish to achieve, for example to add into the schedule a customer order that is left out in the IBS schedule. The algorithm will then decide HOW to accommodate this request on the busy Gantt chart and make the necessary adjustments with minimal degradation to the macro quality scores. More on this in our 2021 blog: A mathematically optimal schedule is not necessarily practically optimal! This blog also discusses the execution time of the schedule and how Plataine Scheduler continues to re-optimize the schedule with its **Context and Actions engine** in response to unexpected disruptions and events.

## 9. A Real-Life Case in Fully Automatic Mode

CarbTech manufactures parts for the aerospace and other industries using composite materials. One of its manufacturing sites involves:

- 75 end products
- 87 stations/machines (including Autoclaves)
- 256 tool types
- 1024 individual tools of all types.

Below are two real-life scenarios illustrating the scale and performance of P-O In FULLY AUTOMATIC mode:

### Scenario 1:

Schedule 1 comprises a Demand Set of 395 work orders over a time horizon of one month. This translates into **7,505 tasks**. With one click on the P-O AutoSchedule button, an optimal Gantt chart was automatically generated in just **1.5 hours.** Considering that before deploying the P-O Scheduler, it could require a hundred or more man-hours by a skilled 3-5 person team to produce a usable Gantt chart, 1.5 hours by P-O is certainly a good practical speed. Add to the speed the great quality metrics and full compliance with all requirements and constraints, and the business value is immediately apparent.

On the same site, we ran a larger-scale scenario as follows:

### Scenario 2:

Schedule 2 comprises a Demand Set of 600 work orders over a time horizon of **3 months**, which translates into **33,548 task**s. With one click on the AutoSchedule button, P-O automatically generated an optimal Gantt chart in 2:45 hours.

### Results

As scenario 2 shows, while the scale of the Demand Set grew substantially—by a factor of 4.5—the computing time in fully automatic mode did not grow in proportion, increasing from 1.5 hours to just 2:45 hours. This is certainly reasonable compared to the time it takes with the alternatives.

Many other scenarios of various complexity were run, and they all confirmed this finding. Note that parallelizing the VHE digital agents still allows for further improvements in computing time.

## 10. Discussion on Bio-inspired Colonies versus "Human Expert Colonies"

**A point to consider:** Let's compare the widely popular bio-inspired algorithms with P-O "human-inspired" algorithms at the fundamental level. On one side, bio-inspired algorithms are based on the intelligence of an Ant Colony, Grasshopper, Fish Swarm, or Bee Colony. On the other side is the new P-O algorithm that aggregates the intelligence of a "colony" of **human experts**, enhanced by reinforced machine learning (RL) that aims to fix individual human experts' weaknesses in specific situations and to combine the collective wisdom to avoid such weaknesses. Each VHE algorithm has evolved over many years of work by many experts. For a given factory's operations and every single Demand Set, the RL machine learning algorithm attempts to overcome the weaknesses of each VHE for this specific case. While respecting the power of millions of years of **animal evolution** leading to the current intelligence of Ants, Grasshoppers, and Fishes and Bees, I would not underestimate the power of each VHE algorithm to begin with and its enhancements by the group/colony wisdom of all the VHEs combined.

The P-O algorithm is based on the deep domain knowledge we acquired from years of working with experienced schedulers and a long list of research and industry publications. Also note that each period brings its unique Demand Sets for the same manufacturing organization, which may differ substantially from what came before. It is quite common that one specific VHE orientation will perform the best for some weeks, while for the Demand Sets of other weeks, it could rank quite low relative to the other VHEs.

## 11. Summary

The Practimum-Optimum™ (P-O) algorithm represents a paradigm shift in developing automatic optimization products for complex real-life business problems such as large-scale manufacturing scheduling. It leverages deep business domain expertise to create a group of virtual human expert (VHE) agents with different 'schools of thought' on how to create high-quality schedules. By computerizing them into algorithms, P-O generates many valid schedules at far higher speeds than human schedulers are capable of. Initially, these schedules can also be local optimum peaks far away from high-quality schedules. By submitting these schedules to a reinforced machine learning algorithm (RL), P-O learns the weaknesses and strengths of each VHE schedule, and accordingly derives 'reward and punishment' changes in the Demand Set that will modify the relative priorities for time and resource allocation that jobs received in the prior iteration that led to the current state of the schedule. These cause the core logic of the VHE algorithms to explore, in the subsequent iteration, substantially different parts of the schedules universe and potentially find higher-quality schedules. Using the hill climbing analogy, this may be viewed as a big jump, shifting from a given local peak to a faraway promising start point equipped with knowledge embedded in the demand set for future iterations.  This is a fundamental difference from most contemporary algorithms, which spend considerable time on local micro- steps restricted to the neighborhoods of local peaks they visit. This difference enables a breakthrough in scale and performance for fully automatic manufacturing scheduling in complex organizations. The P-O algorithm is at the heart of Plataine's Scheduler that, in one click, routinely schedules fully automatically 30,000-50,000 tasks for real-life complex manufacturing operations.

------------------------------------------------------------------------

**Biography of Professor Moshe BenBassat:** For several decades, Prof. BenBassat has been researching, practicing, and educating Artificial Intelligence. During a long academic career with positions at Tel Aviv University, USC, and UCLA, Professor BenBassat made significant contributions to Artificial Intelligence, Optimization, Pattern Recognition, Data Science, and Machine Learning. Following his invention of "service chain optimization" (patent awarded), he founded ClickSoftware Inc., which developed ClickSchedule, the leading AI-based scheduler for field service operations, used worldwide by most of the largest service companies. ClickSoftware is now part of Salesforce.
Prof. BenBassat also founded Plataine Inc., which is focused on intelligent automation for smart manufacturing, developing and selling a suite of enterprise AI applications for decision optimization in manufacturing operations.